\documentclass[twoside,11pt]{article}

\usepackage{jmlr2e}

\firstpageno{1}

\begin{document}

\title{Exploring Swedish \& English fastText Embeddings for NER with the Transformer}

\author{\name Tosin P. Adewumi, Foteini Liwicki \& Marcus Liwicki\\ \email firstname.lastname@ltu.se \\
       \addr EISLAB\\ SRT Department\\
       Luleå University of Technology\\
       Sweden
       }


\maketitle

\begin{abstract}
In this paper, our main contributions are that embeddings from relatively smaller corpora can outperform ones from larger corpora and we make the new Swedish analogy test set publicly available.
To achieve a good network performance in natural language processing (NLP) downstream tasks, several factors play important roles: dataset size, the right hyper-parameters, and well-trained embeddings.
We show that, with the right set of hyper-parameters, good network performance can be reached even on smaller datasets.
We evaluate the embeddings at both the intrinsic and extrinsic levels.
The embeddings are deployed with the Transformer in named entity recognition (NER) task and significance tests conducted.
This is done for both Swedish and English.
We obtain better performance in both languages on the downstream task with smaller training data, compared to recently released, Common Crawl versions; and character n-grams appear useful for Swedish, a morphologically rich language.
\end{abstract}

\begin{keywords}
  Embeddings, Transformer, Analogy, Dataset, NER, Swedish
\end{keywords}

\section{Introduction}
The embedding layer of neural networks may be initialized randomly or replaced with pre-trained vectors, which act as lookup tables.
One of such pre-trained vector tools include fastText, introduced by Joulin et al.
\cite{joulin2016bag}.
The main advantages of fastText are speed and competitive performance to state-of-the-art (SotA).
Using pre-trained embeddings in deep networks like the Transformer can improve performance.
Vaswani et al. (2017) introduced the Transformer, a SotA architecture based on self-attention mechanisms only, and it demonstrated better performance while requiring less time to train \cite{vaswani2017attention}.
Usually, downstream tasks are applied after pre-training language models on such deep networks \cite{brown2020language,devlin2018bert}.

Despite the plethora of embeddings in many languages, there's a dearth of analogy test sets to evaluate many of them, including for Swedish \cite{al2013polyglot,fallgren2016towards,precenth2019word,venekoski2017finnish}.
This is because creating labelled or structured datasets can be expensive in terms of time and attention required. Grave et al. (2018) created 157 different language embeddings but provided analogy test set for only 3 languages: French, Hindi and Polish \cite{grave2018learning}.
An analogy test set, introduced by Mikolov et al. (2013), provides some inclination as to the quality and likely performance of word embeddings in NLP downstream tasks, such as NER, which is used in this work \cite{mikolov2013efficient}.
The evaluation involves prediction of the second value of a pair of two similar words.

Therefore, key contributions of this work (from its objective) are (i) the new Swedish analogy test set publicly made available\footnote{ github.com/tosingithub/tdesk} for the NLP research community, (ii) optimal English and Swedish embeddings, and (iii) insight into their performance in the NER downstream task.
The quality of the Swedish model by Grave et al. (2018) is evaluated, in a first.
The embedding hyper-parameters are based on previous research, which used grid search to determine optimal hyper-parameters \cite{adewumi2020word2vec}.
The rest of this paper is organised as follows: a brief survey of related work, the methodology used, results and discussion, and the conclusion.
 
\section{Related Work}
Distributed representation of words has been around for some time \cite{hinton1986learning}.
fastText, based on the original distributed representation by Mikolov et al. (2013), contains two architectures \cite{mikolov2013efficient}.
Its continuous bag of words (CBoW) averages word vectors into text representation, fed into a linear classifier, while the skipgram uses bag of character n-grams for represented words by summing them \cite{bojanowski2017enriching,joulin2016bag}.
The use of subword representations has proven to be helpful when dealing with out-of-vocabulary (OOV) words.
Indeed, Thomason et al. (2020) used word  embeddings to guide the parsing of OOV words in their work on meaning representation for robots  \cite{thomason2020jointly} .
Swedish, which is close to Finnish, has words that can have many inflected forms and some words may not be present in a corpus \cite{bojanowski2017enriching,fallgren2016towards}.

Despite the potential advantage of subword vectors, Bojanowski et al (2017) observed that using character n-gram was less useful for English compared to some other languages they had explored after a few of the languages were evaluated using different datasets \cite{bojanowski2017enriching}.
It is doubtful if comparison of their results obtained across languages is truly justified, given that different Wikipedia corpora, possibly of different sizes, were trained and tested on different analogy datasets.
This risk was observed by other researchers while working with English and German embeddings, for which they took measures \cite{koper2015multilingual}.

WordSimilarity-353 (WordSim) test set is another analysis tool for word vectors \cite{finkelstein2002placing}.
It is based on human expert-assigned semantic similarity on two sets of English word pairs.
This is unlike analogy score, based on vector space algebra.
Both are used to measure intrinsic embedding quality.
Despite their weaknesses, they have been shown to reveal somewhat meaningful relationships among words in embeddings \cite{mikolov2013efficient,pennington2014glove}.
It is misleading to assume such intrinsic tests are sufficient in themselves, just as it is misleading to assume one particular extrinsic test is sufficient to generalise the performance of embeddings on all NLP tasks \cite{gatt2018survey,faruqui2016problems,adewumi2020word2vec}.
For Swedish, a common evaluation resource for words is SALDO \cite{borin2013saldo}, which is a lexical-semantic resource that links words by their associations.
SALDO extends SAL (Svenskt associationslexikon, a set of classified synonyms) with inflectional morphological information \cite{borin2013saldo,eide2016swedish}.
QVEC-CCA may be used as an intrinsic evaluation metric with features from language resource like SALDO \cite{tsvetkov2016correlation,fallgren2016towards}.

Joulin et al. (2016) noted that other implementations of their fastText model could be much slower \cite{joulin2016bag}.
Indeed, implementations in Python, an interpreted language, are expected to be slower and will use up more energy resources, compared to the original C++ implementation \cite{adewumi2018inner,adewumi2020inner}.
The English and Swedish language models by Grave et al. (2018) were trained on Common Crawl \& Wikipedia datasets, using CBoW of 300 dimensions, with character n-grams of length 5 and window size 5 \cite{grave2018learning}.
These are the embeddings we compare with in this work.
Common Crawl contains petabytes of data, resulting in 630 billion words after preprocessing in a previous use \cite{mikolov2017advances}.

The Transformer, in its original form, maintains an encoder-decoder architecture \cite{vaswani2017attention}.
An input sequence is mapped to a sequence of continuous representations by the encoder.
Then, the decoder makes auto-regressive output sequence of symbols, one at a time, utilizing the previously generated symbols as extra input for the next.
Self-attention, in neural networks, computes a representation of various positions of a sequence and this is what the Transformer architecture employs \cite{vaswani2017attention}.
The Transformer architecture, in one form or the other, has been utilized in recent SotA results  \cite{devlin2018bert,brown2020language}.

\section{Methodology}
\subsection{Upstream}
All pre-trained models in English and Swedish were generated using the original C++ implementation \cite{grave2018learning}.
This forestalls using any sub-optimal, third-party implementations.
They were run on a shared DGX cluster running Ubuntu 18 with 80 CPUs.
Gensim Python library program was used to evaluate all models against their corresponding analogy test sets.
Some of the default hyper-parameter settings were retained \cite{bojanowski2017enriching}.
All models are 300 dimensions and trained for 10 epochs.
The lower and upper boundaries for the character n-gram were 3 and 6, respectivley.
Table 1 identifies other hyper-parameters (and notations used in subsequent tables).

Both the English and Swedish training datasets used are 2019 Wikipedia dumps of 27G (4.86B words) and 4G (767M words), respectively, after pre-processing \cite{enwiki,svwiki}.
They were pre-processed using the recommended script \cite{grave2018learning}.
It would have been ideal to run each training multiple times to obtain averages but because of the limited time involved, a work-around was adopted, which was to run a few random models twice to ascertain if there were major differences per model.
It was established that differences were little enough to accept a single run per model.
Besides, each run took hours within the range of about 2 and 36 hours and there were 32 pre-trained models to be generated: 8 English subword and no-subword (word2vec) models each and 8 Swedish subword and no-subword models each.

\begin{table}[ht]
\centering
\begin{tabular}{c|c}
\textbf{Hyper-parameter} & \textbf{Values} \\
\hline
Window size (w) &  4, 8 \\
\hline
Architecture & Skipgram (s1), CBoW (s0) \\
\hline
Loss Function & H. Softmax (h1), N. Sampling (h0)\\
\hline
\end{tabular}
\label{hyper}
\caption{Hyper-parameter choices}
\end{table}

\subsection{Downstream}
The downstream tasks were run on the same cluster mentioned earlier but on Tesla V100 GPU.
The models and source codes are available\textsuperscript{1}.
Selected pre-trained embeddings were evaluated, for both languages, using the Transformer Encoder architecture in PyTorch.
This is without language model pre-training of the Transformer.
There are other models/architectures that can be applied to NER, such as conditional random field (CRF)-based models \citep{finkel2005incorporating, manning2014stanford} but this can be left to future work.
Two corpora were used for the NER downstream task:
Groningen Meaning Bank (GMB) for the Englsih NER \cite{bos2017groningen} and the Stockholm Internet Corpus (SiC) \cite{sic2017}.
GMB contains 47,959 sentence samples, with 17 labels from 9 main labels and 2 context tags.
SiC contains 13,562 samples and follows the CoNLL \& SUC 3.0 (Stockholm-Umeå Corpus) formats.
It has 3 main tags and 8 types, resulting in 17 possible label combinations, however, in practice, 14 labels are currently represented in the corpus.

In both language cases of the NER experiments, the default PyTorch embedding was tested before being replaced by the pre-trained embeddings, with frozen weights.
In each case, the dataset was shuffled before training and split in the ratio 70:15:15 for training, dev and test sets.
Three hyper-parameters were tuned using SigOpt (Bayesian hyper-parameter optimization tool) for 45 combinations (or observation budget) over the network optimizer (between Adam \& RMSProp), Transformer layers (6-12) and attention heads (2-6) \cite{pmlr-v84-martinez-cantin18a}.
This approach eliminates the need to explore all possible combinations in a grid search.
For the English NER, SigOpt optimized and reported the following values: 7 layers, 3 heads and Adam optimizer.
These values were then kept constant for all other embeddings in English.
The same was done for the Swedish NER after optimized values obtained were 8 layers, 2 heads and RMSProp optimizer.
Batch size of 64 was used and each experiment conducted five times and average values reported.
Each run of experiment was for 20 epochs.
However, after validation at each epoch, the model is saved, if it has lower loss than a previous value, thereby avoiding overfitting.
The saved model is then used to evaluate the test set.

\subsection{Swedish analogy test set}
The Swedish analogy test set follows the format of the original Google version.
The original has been observed to be slightly unbalanced, having 8,869 semantic samples and 10,675 syntactic samples (making a total of 19,544).
The Swedish set is bigger and balanced across the 2 major categories, having a total of 20,637, made up of 10,380 semantic and 10,257 syntactic samples.
It is also roughly balanced across the syntactic subsections but the \textit{capital-world} has the largest proportion of samples in the semantic subsection.
This is because of the difficulty involved in obtaining world currencies in Swedish and the limited nomenclature of family members.
A similar difficulty was experienced by Venekoski \& Vankka (2017), who noted that not all words in the original Google analogy test set can be directly translated to other languages, while creating a much smaller Finnish version.
In all, there are 5 semantic subsections and 6 syntactic subsections.
Table 2 presents further details on the test set.
It was constructed, partly using the samples in the English version, with the help of tools dedicated to Swedish dictionary/translation\footnote{https://bab.la \& https://en.wiktionary.org/wiki/} and was proof-read for corrections by two native speakers (with a percentage agreement of 98.93\%).
New, relevant entries were also added.
The famous sample in the family subsection of the semantic section is: \textit{kung drottning man kvinna}.

\begin{table}[ht]
\centering
\begin{tabular}{c|c}
\textbf{Semantic} & \textbf{Syntactic} \\
\hline
\footnotesize{capital-common-countries (342)} & \footnotesize{gram2-opposite (2,652)} \\
\hline
\footnotesize{capital-world (7,832)} &
\footnotesize{gram3-comparative (2,162)} 
\\
\hline
\footnotesize{currency (42)} & 
\footnotesize{gram4-superlative (1,980)}
\\
\hline
\footnotesize{city-in-state (1,892)} & 
\footnotesize{gram6-nationality-adjective (12)} 
\\
\hline
\footnotesize{family (272)} &
\footnotesize{gram7-past-tense (1,891)}
\\
\hline
 & 
\footnotesize{gram8-plural (1,560)}
\\
\hline
\end{tabular}
\label{svtest}
\caption{Swedish analogy test set details}
\end{table}

\section{Results \& Discussion}
The WordSim result output file from the Gensim Python program always has more than one value reported, including the Spearman correlation.
The first value is reported as WordSim score1 in the relevant table.
Intrinsic results for the pre-trained models are given in table 3.
An important trend that can be observed is the higher scores for skipgram-negative sampling in all the cases (English \& Swedish), except one.
This appears to confirm previous research \cite{mikolov2013efficient,adewumi2020word2vec}.
It is noteworthy that the released, original pre-trained word2vec model was of the same combination \cite{mikolov2013efficient}.
This English word2vec (no-subword) embedding was trained on GoogleNews dataset of 100 billion words and represented as \textit{'GN'} in the table \cite{mikolov2013efficient}.
The English subword embeddings have 5 models with higher analogy scores than their word2vec equivalent, out of 8.
The WordSim and corresponding Spearman correlation for English word2vec models were higher than their corresponding subword models in all cases, except one.
It may not be proper to compare the scores of the English to the Swedish models since both were based on different test sets of varying sizes.

Given the observation that using character n-gram was less useful for English than some other languages, it's not expected that the scores will follow a similar trend for all languages \cite{bojanowski2017enriching}.
In addition, accuracy falls for morphologically complex languages, like German, making analogy predictions difficult \cite{koper2015multilingual}.
While working on Finnish embeddings, it was observed that fastText (subword) CBoW had lower analogy score than word2vec CBoW while fastText skipgram had higher score than word2vec skipgram, even for zero OOV words \cite{venekoski2017finnish}.

Indeed, determining the best pre-trained model in each category requires the additional step of applying them to downstream tasks, in this case NER \cite{chiu2016intrinsic}.
Tables 4 \& 5 present the results of the NER task for the selected English \& Swedish embeddings, respectively.
The embeddings by Grave et al. (2018), trained on the larger Common Crawl \& Wikipedia, are represented by '\textit{Gr}' in the tables.
It can be observed that for English, the word2vec w8s0h0 embedding outperformed the subword embedding: \textit{Gr}.
The Swedish subword embedding, \textit{Gr}, is also outperformed by the subword embeddings the authors created.
Importantly, the subword versions outperform the word2vec ones, implying the character n-grams may be useful for Swedish.
In both language cases, the good performance of PyTorch default embedding is noticeable.


\begin{table}[ht]
\centering
\begin{tabular}{c|c|c|c|c|c|c|c|c|c|c}
\textbf{} &
\multicolumn{4}{c}{\textbf{Skipgram (s1)}} & \multicolumn{4}{|c|}{\textbf{CBoW (s0)}} &
\textbf{} &
\textbf{} \\
\textbf{} &
\multicolumn{2}{c}{\footnotesize{\textbf{H. S. (h1)}}} &
\multicolumn{2}{|c}{\footnotesize{\textbf{N. S. (h0)}}} &
\multicolumn{2}{|c}{\footnotesize{\textbf{H. S. (h1)}}} &
\multicolumn{2}{|c|}{\footnotesize{\textbf{N. S. (h0)}}} &
\footnotesize{\textbf{Gr}} &
\footnotesize{\textbf{GN}}\\
\hline
\textbf{window (w)} &
\footnotesize{\textbf{4}} & \footnotesize{\textbf{8}} & \footnotesize{\textbf{4}} &
\footnotesize{\textbf{8}} & \footnotesize{\textbf{4}} & \footnotesize{\textbf{8}} & \footnotesize{\textbf{4}} & \footnotesize{\textbf{8}} &
\textbf{} &
\textbf{} \\
\hline
\footnotesize{\textbf{Subword \%}} &
\multicolumn{8}{c}{} &
\textbf{} &
\textbf{} \\
\hline
\footnotesize{Analogy} &
\footnotesize{62.6} & \footnotesize{58.8} & \footnotesize{74.4} & \footnotesize{69.8} & \footnotesize{67.2} & \footnotesize{68.7} & \footnotesize{71.6} & \footnotesize{71} &
\footnotesize{\textbf{82.6}} &
{} \\
\hline
\footnotesize{WordSim score1} &
\footnotesize{64.8} & \footnotesize{66.3} & \footnotesize{69.9} & \footnotesize{\textbf{70}} & \footnotesize{62.6} & \footnotesize{66.2} & \footnotesize{47.3} & \footnotesize{51.1} &
\footnotesize{68.5} &
{} \\
\hline
\footnotesize{Spearman} &
\footnotesize{67.6} & \footnotesize{69.4} & \footnotesize{\textbf{74.3}} & 
\footnotesize{73.6} & \footnotesize{65.3} & \footnotesize{70.3} & \footnotesize{45.3} & \footnotesize{49.5} &
\footnotesize{70.2} &
{} \\
\hline
\footnotesize{\textbf{Word2Vec \%}} &
\multicolumn{8}{c}{} &
\textbf{} &
\textbf{} \\
\hline
\footnotesize{Analogy} &
\footnotesize{61.3} & \footnotesize{58.3} & \footnotesize{73.5} & \footnotesize{70.4} & \footnotesize{59.7} & \footnotesize{61.9} & \footnotesize{\textbf{76.2}} & \footnotesize{75.4} &
{} &
\footnotesize{74} \\
\hline
\footnotesize{WordSim score1} &
\footnotesize{66.3} & \footnotesize{67.3} & \footnotesize{69.6} & \footnotesize{\textbf{70.1}} & \footnotesize{64.1} & \footnotesize{66.7} & \footnotesize{65.4} & \footnotesize{67.5} &
{} &
\footnotesize{62.4} \\
\hline
\footnotesize{Spearman} &
\footnotesize{70} & \footnotesize{70.9} & \footnotesize{74.5} & \footnotesize{\textbf{74.7}} & \footnotesize{68.2} & \footnotesize{71.2} & \footnotesize{66.9} & \footnotesize{69.4} &
{} &
\footnotesize{65.9} \\
\hline
\multicolumn{11}{|c|}{\footnotesize{\textbf{Swedish}}} \\
\hline
\footnotesize{\textbf{Subword \%}} &
\footnotesize{45.05} & \footnotesize{39.99} & \footnotesize{53.53} & \footnotesize{53.36} & \footnotesize{26.5} & \footnotesize{23.93} & \footnotesize{36.79} & \footnotesize{35.89} &
\footnotesize{\textbf{60.9}} &
\footnotesize{} \\
\hline
\footnotesize{\textbf{Word2Vec \%}} &
\footnotesize{45.53} & \footnotesize{41.21} & \footnotesize{\textbf{58.25}} & \footnotesize{57.30} & \footnotesize{28.02} & \footnotesize{28.04} & \footnotesize{52.81} & \footnotesize{55.64} &
{} &
\footnotesize{} \\
\hline
\end{tabular}
\label{subwords_scores}
\caption{Intrinsic Scores - English \& Swedish (highest score/row in bold)}
\end{table}

Significance tests, using bootstrap \cite{calmettes2012making}, on the results of the differences in means of the English \textit{Gr} \& word2vec w8s0h0 models, show a 95\% confidence interval (CI) of [0.0003, 0.1674] but [-0.3257,  0.169] for Swedish \textit{Gr} \& subword w4s1h1.
The CI interval for English does not include 0, though the lower limit is small, thus we can conclude the difference is unlikely due to chance but the CI for Swedish includes 0, thus the difference is likely due to chance.

\begin{table}[ht]
\centering
\resizebox{\columnwidth}{!}{%
\begin{tabular}{c|c|c|c|c|c|c|c|c|c|c|c|c|c|c}
\footnotesize{} &
\multicolumn{2}{c}{\footnotesize{}} & \multicolumn{2}{c}{\footnotesize{}} & 
\multicolumn{6}{|c|}{\footnotesize{\textbf{Word2Vec (W)}}} & 
\multicolumn{4}{c}{\footnotesize{\textbf{Subword}}}
\\
\hline
\footnotesize{\textbf{Metric}} &
\multicolumn{2}{c|}{\footnotesize{\textbf{Default}}} & \multicolumn{2}{c|}{\footnotesize{\textbf{Gr}}} & 
\multicolumn{2}{c|}{\footnotesize{\textbf{w8s0h0}}} & \multicolumn{2}{c|}{\footnotesize{\textbf{w4s0h0}}} &
\multicolumn{2}{c|}{\footnotesize{\textbf{w4s1h0}}} &
\multicolumn{2}{c|}{\footnotesize{\textbf{w4s0h0}}} & 
\multicolumn{2}{c}{\footnotesize{\textbf{w8s1h1}}}\\
\hline
\footnotesize{} &
\footnotesize{Dev} &
\footnotesize{Test} & \footnotesize{Dev} &
\footnotesize{Test} & \footnotesize{Dev} &
\footnotesize{Test} &
\footnotesize{Dev} &
\footnotesize{Test} & \footnotesize{Dev} &
\footnotesize{Test} &
\footnotesize{Dev} &
\footnotesize{Test} &
\footnotesize{Dev} &
\footnotesize{Test}\\
\hline
\footnotesize{F1} &
\footnotesize{0.719} &
\footnotesize{\textbf{0.723}} & \footnotesize{0.588} &
\footnotesize{0.6602} & \footnotesize{0.719} &
\footnotesize{0.720} &
\footnotesize{0.715} &
\footnotesize{0.716} & \footnotesize{0.714} &
\footnotesize{0.716} &
\footnotesize{0.695} &
\footnotesize{0.668} &
\footnotesize{0.592} &
\footnotesize{0.684}\\
\hline
\footnotesize{Precision} &
\footnotesize{0.685} &
\footnotesize{0.69} & \footnotesize{0.564} &
\footnotesize{0.634} & \footnotesize{0.689} &
\footnotesize{\textbf{0.691}} &
\footnotesize{0.686} &
\footnotesize{0.688} & \footnotesize{0.684} &
\footnotesize{0.686} &
\footnotesize{0.664} &
\footnotesize{0.64} &
\footnotesize{0.567} &
\footnotesize{0.656}\\
\hline
\footnotesize{Recall} &
\footnotesize{0.756} &
\footnotesize{\textbf{0.759}} & \footnotesize{0.615} &
\footnotesize{0.689} & \footnotesize{0.751} &
\footnotesize{0.752} &
\footnotesize{0.747} &
\footnotesize{0.747} & \footnotesize{0.748} &
\footnotesize{0.748} &
\footnotesize{0.729} &
\footnotesize{0.7} &
\footnotesize{0.62} &
\footnotesize{0.713}\\
\hline
\end{tabular}
}
\label{tabner_eng}
\caption{English NER Mean Scores}
\end{table}

\begin{table}[ht]
\centering
\resizebox{\columnwidth}{!}{%
\begin{tabular}{c|c|c|c|c|c|c|c|c|c|c|c|c|c|c}
\footnotesize{} &
\multicolumn{2}{c}{\footnotesize{}} & \multicolumn{2}{c}{\footnotesize{}} & 
\multicolumn{4}{|c|}{\footnotesize{\textbf{Word2Vec (W)}}} & 
\multicolumn{6}{c}{\footnotesize{\textbf{Subword}}}
\\
\hline
\footnotesize{\textbf{Metric}} &
\multicolumn{2}{c|}{\footnotesize{\textbf{Default}}} & \multicolumn{2}{c|}{\footnotesize{\textbf{Gr}}} & 
\multicolumn{2}{c|}{\footnotesize{\textbf{w4s1h0}}} & 
\multicolumn{2}{c|}{\footnotesize{\textbf{w8s0h1}}} &
\multicolumn{2}{c|}{\footnotesize{\textbf{w4s1h1}}} & \multicolumn{2}{c|}{\footnotesize{\textbf{w4s1h0}}} &
\multicolumn{2}{c}{\footnotesize{\textbf{w8s0h1}}}
\\
\hline
\footnotesize{} &
\footnotesize{Dev} &
\footnotesize{Test} & \footnotesize{Dev} &
\footnotesize{Test} & \footnotesize{Dev} &
\footnotesize{Test} &
\footnotesize{Dev} &
\footnotesize{Test} & \footnotesize{Dev} &
\footnotesize{Test} &
\footnotesize{Dev} &
\footnotesize{Test} &
\footnotesize{Dev} &
\footnotesize{Test}\\
\hline
\footnotesize{F1} &
\footnotesize{0.487} &
\footnotesize{\textbf{0.675}} & \footnotesize{0.441} &
\footnotesize{0.568} &
\footnotesize{0.574} &
\footnotesize{0.344} &
\footnotesize{0.477} &
\footnotesize{0.429} &
\footnotesize{0.507} &
\footnotesize{0.649} &
\footnotesize{0.492} &
\footnotesize{0.591} & \footnotesize{0.486} &
\footnotesize{0.623}
\\
\hline
\footnotesize{Precision} &
\footnotesize{0.51} &
\footnotesize{0.745} & \footnotesize{0.682} &
\footnotesize{\textbf{0.856}} &
\footnotesize{0.704} &
\footnotesize{0.549} &
\footnotesize{0.626} &
\footnotesize{0.669} & \footnotesize{0.647} &
\footnotesize{0.821} &
\footnotesize{0.658} &
\footnotesize{0.752} & \footnotesize{0.626} &
\footnotesize{0.802}
\\
\hline
\footnotesize{Recall} &
\footnotesize{0.471} &
\footnotesize{\textbf{0.633}} & \footnotesize{0.331} &
\footnotesize{0.44} &
\footnotesize{0.489} &
\footnotesize{0.265} &
\footnotesize{0.398} &
\footnotesize{0.325} & \footnotesize{0.420} &
\footnotesize{0.543} &
\footnotesize{0.398} &
\footnotesize{0.5} & \footnotesize{0.402} &
\footnotesize{0.524}
\\
\hline
\end{tabular}
}
\label{tabner_sv}
\caption{Swedish NER Mean Scores}
\end{table}

\begin{figure}[!htb]
   \begin{minipage}{1\textwidth}
     \centering
     \includegraphics[width=.9\linewidth]{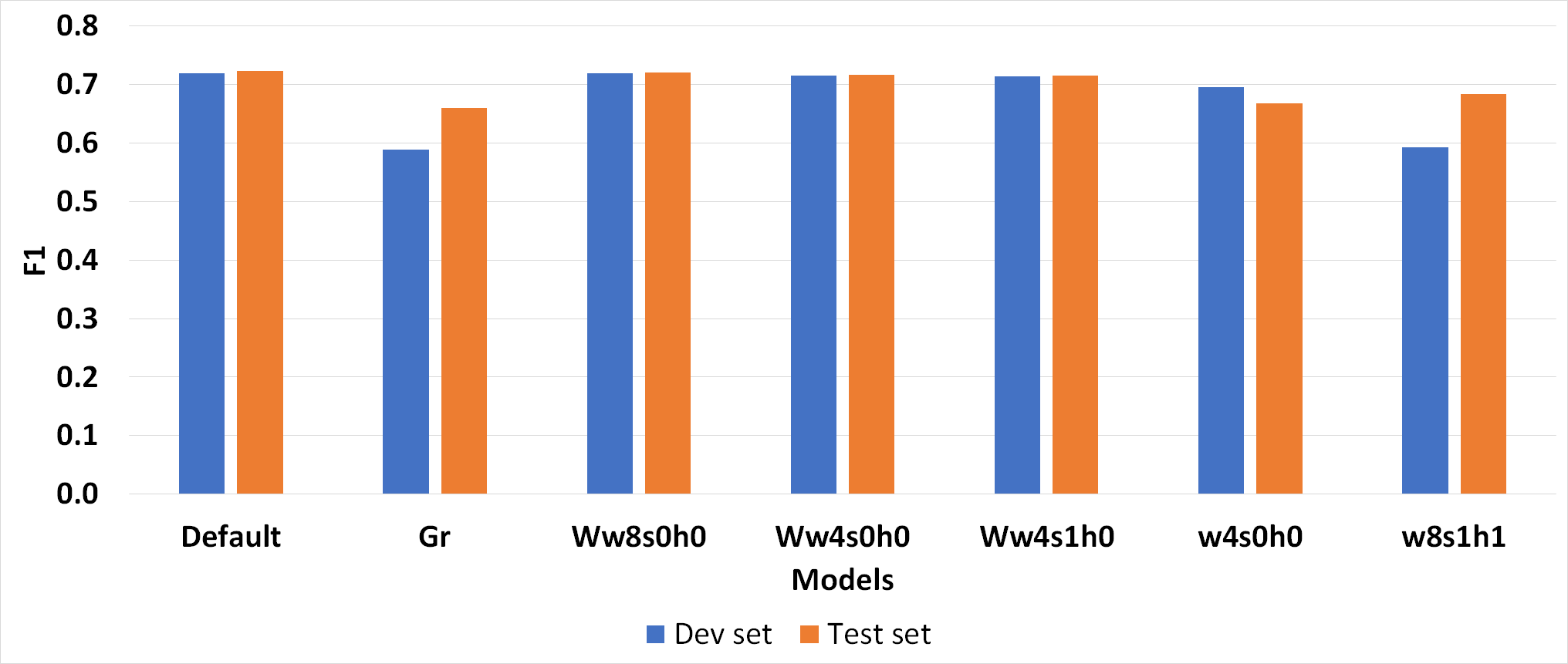}
     \caption{English NER mean F1 scores}\label{FigengNER}
   \end{minipage}\hfill
   \begin{minipage}{1\textwidth}
     \centering
     \includegraphics[width=.9\linewidth]{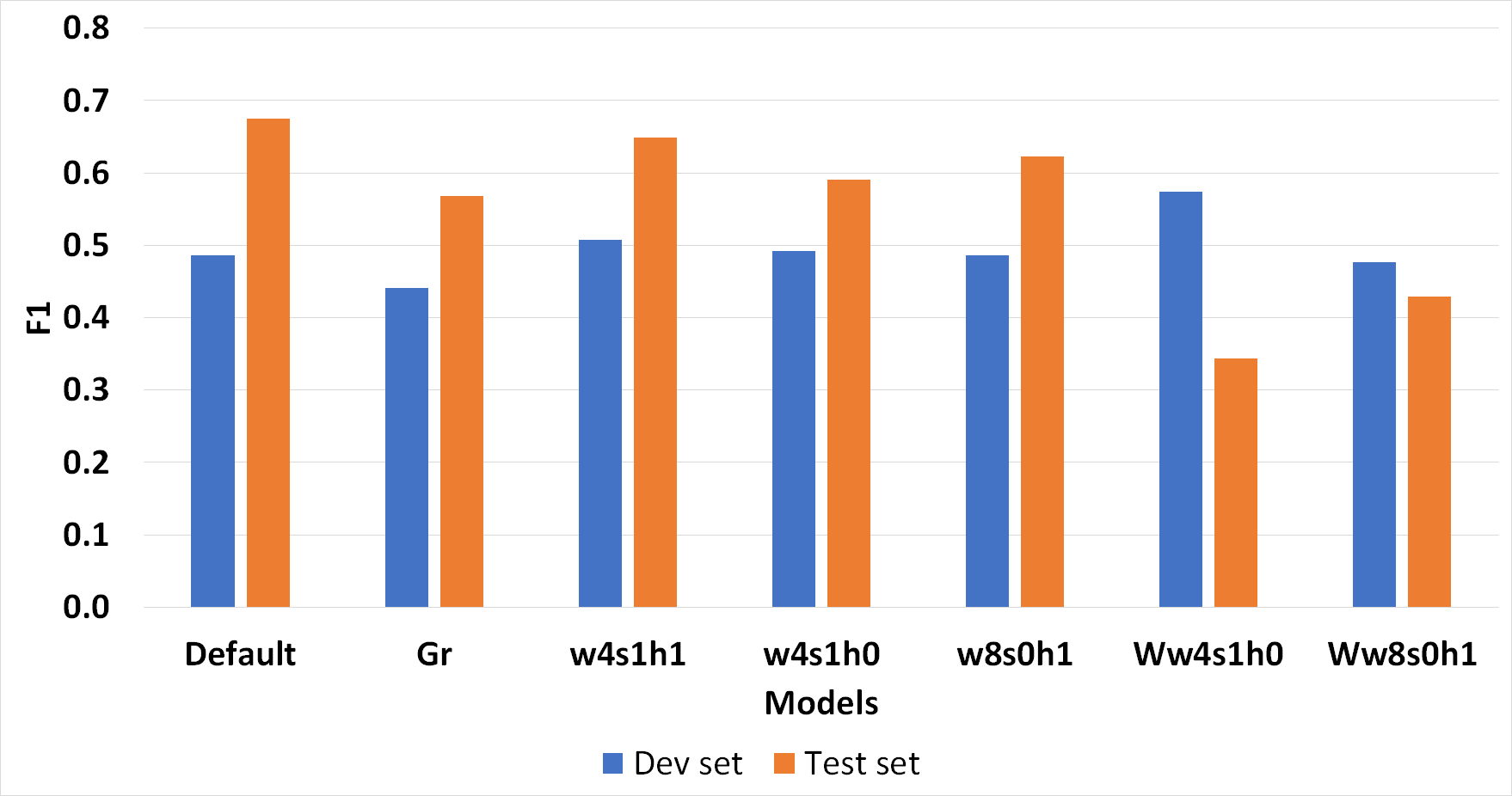}
     \caption{Swedish NER mean F1 scores}\label{FigsweNER}
   \end{minipage}
\end{figure}

\newpage
\subsection{Embedding Qualitative Assessment}
Qualitative assessment of the Swedish model (subword w4s1h1) in one instance is given in table 6, for randomly selected input.

\begin{table}[ht]
\centering
\begin{tabular}{c|c}
\textbf{\footnotesize{Nearest Neighbor/ Analogy Query}} & \textbf{\footnotesize{Result}} \\
\hline
\footnotesize{syster} & \footnotesize{halvsyster (0.8688), systerdotter (0.8599), ...} \\
\hline
\footnotesize{rom - italien + kairo} & \footnotesize{egypten (0.4889), norditalien (0.4317), ...}\\
\hline
\end{tabular}
\label{qualityembedding}
\caption{Qualitative assessment of Swedish w4s1h1 model}
\end{table}

\subsection{Learning Qualitative Assessment}
It was observed that learning occurs faster with the Transformer than the LSTM, which was used in an earlier work.
Tables 7 \& 8 provide examples for both languages.
In one instance, in the English case, learning almost correctly occurs by epoch 5.
We observed that most times it's earlier.
A similar occurrence is observed with Swedish.
The learning is not always 100\% correct, though.

\begin{table}[ht]
\centering
\resizebox{\columnwidth}{!}{%
\begin{tabular}{c|cccccccccccccc}
\footnotesize{\textbf{}} &
\multicolumn{14}{c}{\footnotesize{\textbf{Sample Sentence Tokens/ Tags}}}\\
\hline
\footnotesize{Sentence:} &
\footnotesize{Turkey} &
\footnotesize{'s} & \footnotesize{Foreign} &
\footnotesize{Ministry} & \footnotesize{says} &
\footnotesize{several} &
\footnotesize{of} &
\footnotesize{its} &
\footnotesize{nationals} &
\footnotesize{were} &
\footnotesize{killed} &
\footnotesize{Friday} &
\footnotesize{in} &
\footnotesize{an}\\
\footnotesize{} &
\footnotesize{ambush} & \footnotesize{in} &
\footnotesize{the} & \footnotesize{northern} &
\footnotesize{Iraqi} &
\footnotesize{city} &
\footnotesize{of} & \footnotesize{Mosul} &
\footnotesize{.} &
\footnotesize{} &
\footnotesize{} &
\footnotesize{} &
\footnotesize{} &
\footnotesize{}\\
\hline
\footnotesize{True Tags} &
\footnotesize{B-org} &
\footnotesize{I-org} & \footnotesize{I-org} &
\footnotesize{I-org} & \footnotesize{O} &
\footnotesize{O} &
\footnotesize{O} &
\footnotesize{O} & \footnotesize{O} &
\footnotesize{O} &
\footnotesize{O} &
\footnotesize{B-tim} &
\footnotesize{O} &
\footnotesize{O}\\
\footnotesize{} &
\footnotesize{O} &
\footnotesize{O} & \footnotesize{O} &
\footnotesize{O} & \footnotesize{B-gpe} &
\footnotesize{O} &
\footnotesize{O} &
\footnotesize{B-geo} & \footnotesize{O} &
\footnotesize{} &
\footnotesize{} &
\footnotesize{} &
\footnotesize{} &
\footnotesize{}\\
\hline
\footnotesize{Tags@Epoch 1} &
\footnotesize{B-geo} &
\footnotesize{O} & \footnotesize{O} &
\footnotesize{O} & \footnotesize{O} &
\footnotesize{O} &
\footnotesize{O} &
\footnotesize{O} & \footnotesize{O} &
\footnotesize{O} &
\footnotesize{O} &
\footnotesize{B-tim} &
\footnotesize{O} &
\footnotesize{O}\\
\footnotesize{} &
\footnotesize{O} &
\footnotesize{O} & \footnotesize{O} &
\footnotesize{O} & \footnotesize{B-gpe} &
\footnotesize{O} &
\footnotesize{O} &
\footnotesize{B-geo} & \footnotesize{O} &
\footnotesize{} &
\footnotesize{} &
\footnotesize{} &
\footnotesize{} &
\footnotesize{}\\
\hline
\footnotesize{Tags@Epoch 2} &
\footnotesize{B-geo} &
\footnotesize{O} & \footnotesize{O} &
\footnotesize{I-org} & \footnotesize{O} &
\footnotesize{O} &
\footnotesize{O} &
\footnotesize{O} & \footnotesize{O} &
\footnotesize{O} &
\footnotesize{O} &
\footnotesize{B-tim} &
\footnotesize{O} &
\footnotesize{O}\\
\footnotesize{} &
\footnotesize{O} &
\footnotesize{O} & \footnotesize{O} &
\footnotesize{O} & \footnotesize{B-gpe} &
\footnotesize{O} &
\footnotesize{O} &
\footnotesize{B-geo} & \footnotesize{O} &
\footnotesize{} &
\footnotesize{} &
\footnotesize{} &
\footnotesize{} &
\footnotesize{}\\
\hline
\footnotesize{Tags@Epoch 5} &
\footnotesize{B-org} &
\footnotesize{O} & \footnotesize{I-org} &
\footnotesize{I-org} & \footnotesize{O} &
\footnotesize{O} &
\footnotesize{O} &
\footnotesize{O} & \footnotesize{O} &
\footnotesize{O} &
\footnotesize{O} &
\footnotesize{B-tim} &
\footnotesize{O} &
\footnotesize{O}\\
\footnotesize{} &
\footnotesize{O} &
\footnotesize{O} & \footnotesize{O} &
\footnotesize{O} & \footnotesize{B-gpe} &
\footnotesize{O} &
\footnotesize{O} &
\footnotesize{B-geo} & \footnotesize{O} &
\footnotesize{} &
\footnotesize{} &
\footnotesize{} &
\footnotesize{} &
\footnotesize{}\\
\hline
\end{tabular}
}
\label{tabengsample}
\caption{English Learning Sample}
\end{table}

\begin{table}[ht]
\centering
\resizebox{\columnwidth}{!}{%
\begin{tabular}{c|cccccccccccccc}
\footnotesize{\textbf{}} &
\multicolumn{14}{c}{\footnotesize{\textbf{Sample Sentence Tokens/ Tags}}}\\
\hline
\footnotesize{Sentence:} &
\footnotesize{Även} &
\footnotesize{kollat} & \footnotesize{upp} &
\footnotesize{lite} & \footnotesize{tågresor} &
\footnotesize{till} &
\footnotesize{Borlänge} &
\footnotesize{i} &
\footnotesize{sommar} &
\footnotesize{!} &
\footnotesize{} &
\footnotesize{} &
\footnotesize{} &
\footnotesize{}\\
\hline
\footnotesize{True Tags} &
\footnotesize{O} &
\footnotesize{O} & \footnotesize{O} &
\footnotesize{O} & \footnotesize{O} &
\footnotesize{O} &
\footnotesize{Bplace} &
\footnotesize{O} & \footnotesize{O} &
\footnotesize{O} &
\footnotesize{} &
\footnotesize{} &
\footnotesize{} &
\footnotesize{}\\
\hline
\footnotesize{Tags@Epoch 1} &
\footnotesize{O} &
\footnotesize{O} & \footnotesize{O} &
\footnotesize{O} & \footnotesize{O} &
\footnotesize{O} &
\footnotesize{O} &
\footnotesize{O} & \footnotesize{O} &
\footnotesize{O} &
\footnotesize{} &
\footnotesize{} &
\footnotesize{} &
\footnotesize{}\\
\hline
\footnotesize{Tags@Epoch 2} &
\footnotesize{O} &
\footnotesize{O} & \footnotesize{O} &
\footnotesize{O} & \footnotesize{O} &
\footnotesize{O} &
\footnotesize{Bplace} &
\footnotesize{O} & \footnotesize{O} &
\footnotesize{O} &
\footnotesize{} &
\footnotesize{} &
\footnotesize{} &
\footnotesize{}\\
\hline
\end{tabular}
}
\label{tabsvsample}
\caption{Swedish Learning Sample}
\end{table}

\section{Conclusion}
This work has presented optimal fastText embeddings in Swedish and English for NLP purposes.
It has also presented the first Swedish analogy test set for intrinsic evaluation of Swedish embeddings.
The intrinsic evaluation shows the trend of better performance with skipgram-negative sampling pre-trained models across the two languages.
We also observe that for downstream evaluation for English, the word2vec embedding: CBoW-negative sampling of window size 8, like its other counterparts, outperform the subword embedding of the bigger Common Crawl dataset.
From the results, it may be that WordSim makes better predictions of the performance on downstream tasks.
The Swedish subword embeddings outperform the word2vec versions, implying that character n-grams may be useful for Swedish, a morphologically rich language.
Also, they outperform the subword embedding of the larger Common Crawl dataset.

Merely increasing training dataset size does not equate to better performance and optimal hyper-parameters can improve performance \cite{adewumi2020word2vec}.
Future work can evaluate embeddings of language model pre-training of the Transformer-based SotA models and other downstream tasks.
Other Machine Learning frameworks may also be evaluated.

\acks{The authors wish to thank Carl Borngrund and Karl Ekström for their very useful help in proof-reading the analogy set.
The work in this project is partially funded by Vinnova under the project number 2019-02996 "Språkmodeller för svenska myndigheter".}







\vskip 0.2in
\bibliography{sample}

\end{document}